\documentclass[10pt,twocolumn,letterpaper]{article}

\usepackage{iccv}
\usepackage{times}
\usepackage{epsfig}
\usepackage{graphicx}
\usepackage{amsmath}
\usepackage{amssymb}
\usepackage{multirow}
\usepackage{xcolor}
\usepackage{colortbl}
\usepackage{pifont}

\usepackage[pagebackref=true,breaklinks=true,letterpaper=true,colorlinks,bookmarks=false]{hyperref}

\iccvfinalcopy 

\newcommand{\tablestyle}[2]{\setlength{\tabcolsep}{#1}\renewcommand{\arraystretch}{#2}\centering\footnotesize}

\renewcommand{\paragraph}[1]{\vspace{1.25mm}\noindent\textbf{#1}}

\newcommand{\cmark}{\ding{51}}%
\newcommand{\xmark}{\text{\ding{55}}}

\definecolor{mygray}{gray}{.92}

\newlength\savewidth\newcommand\shline{\noalign{\global\savewidth\arrayrulewidth
  \global\arrayrulewidth 1pt}\hline\noalign{\global\arrayrulewidth\savewidth}}

\def\iccvPaperID{****} 

\ificcvfinal\fi

\begin{document}

\title{POSTER V2: A simpler and stronger facial expression recognition network}

\author{
Jiawei Mao$^\dag$
\quad Rui Xu$^\dag$  \quad Xuesong Yin{\thanks{Corresponding author.$^\dag$Equal contribution.}} \quad Yuanqi Chang \quad Binling Nie \quad Aibin Huang$^*$  \\ 
School of Media and Design, Hangzhou Dianzi University, Hangzhou, China \qquad \\
{\tt\small\{jiaweima0,211330017,yinxs,yuanqichang,binlingnie,huangaibin\}@hdu.edu.cn }\\
}

\maketitle
\ificcvfinal\thispagestyle{empty}\fi

\begin{abstract}
   Facial expression recognition (FER) plays an important role in a variety of real-world applications such as human-computer interaction. 
   POSTER achieves the state-of-the-art (SOTA) performance in FER by effectively combining facial landmark and image features through two-stream pyramid cross-fusion design. 
   However, the architecture of POSTER is undoubtedly complex. It causes expensive computational costs. 
   In order to relieve the computational pressure of POSTER, in this paper, we propose POSTER++. It improves POSTER in three directions: cross-fusion, two-stream, and multi-scale feature extraction. 
   In cross-fusion, we use window-based cross-attention mechanism replacing vanilla cross-attention mechanism. 
   We remove the image-to-landmark branch in the two-stream design. For multi-scale feature extraction, POSTER++ combines images with landmark's multi-scale features to replace POSTER's pyramid design. 
   Extensive experiments on several standard datasets show that our POSTER++ achieves the SOTA FER performance with the minimum computational cost. 
   For example, POSTER++ reached 92.21\% on RAF-DB, 67.49\% on AffectNet (7 cls) and 63.77\% on AffectNet (8 cls), respectively, using only 8.4G floating point operations (FLOPs) and 43.7M parameters (Param). 
   This demonstrates the effectiveness of our improvements.
\end{abstract}

\section{Introduction}

With the continuous development of technology and the continuous improvement of automation, the need for human-computer interaction is becoming increasingly strong. 
Facial expression recognition (FER) helps machines to understand human emotions from facial expressions. 
This makes it as a core task for human-computer interaction. Besides, with its powerful expression understanding ability, FER has great potential applications in psychology, intelligent robotics, intelligent surveillance, virtual reality and synthetic animation. 
Therefore, research on FER is very necessary.

\begin{figure}[t]
   \centering
   \includegraphics[width=1\linewidth]{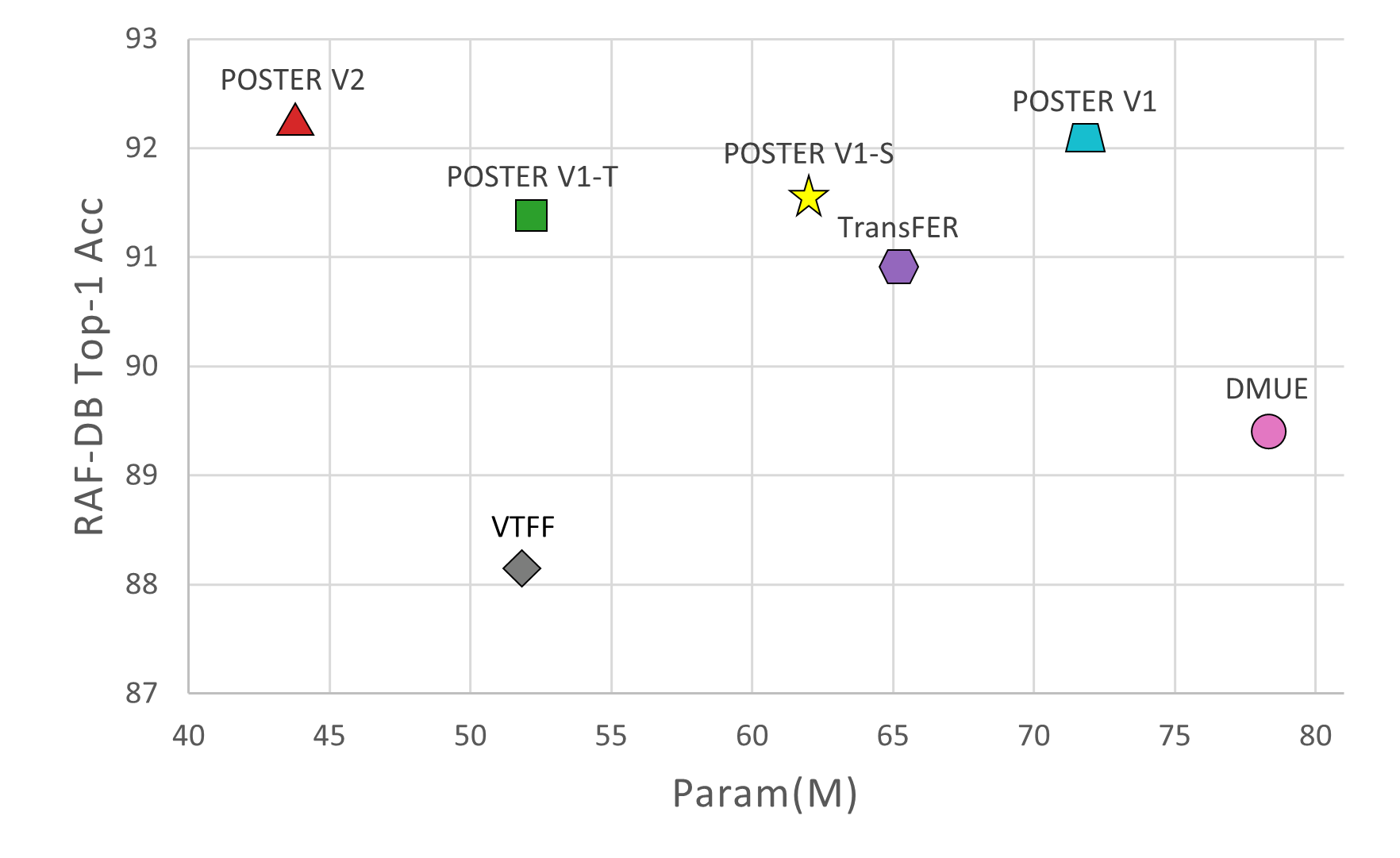}
    \caption{POSTER++ results on RAF-DB. We compare POSTER++ with three variants of POSTER and other FER algorithms. 
    The results indicate that POSTER++ weighs the number of parameters and accuracy better than other FER methods on RAF-DB.}
    \label{fig1}
 \end{figure}

Due to the increasing attention of FER, it has been able to develop rapidly in recent years. 
Early FER works \cite{zhao2007dynamic,zhong2012learning,moore2011local,hu2008multi} used manual features \cite{dalal2005histograms,ojala2002multiresolution,jabid2010local} for the analysis of facial expressions. 
However, FER algorithms based on manual features are often only applicable to specific FER tasks. 
When applied to real world scenarios, it is difficult for these algorithms to achieve the same results as in the experimental setting. 
With the development of deep learning, convolutional neural networks (CNNs) are introduced to FER for improving the robustness of the network. 
Savchenko et al. \cite{savchenko2021facial} first verified the effectiveness of CNNs such as MobileNet \cite{howard2017mobilenets}, EfficientNet \cite{tan2019efficientnet} and RexNet \cite{han2020rexnet} for FER. 
Zhao et al. proposed an efficient and robust FER network EfficientFace \cite{zhao2021robust} for the analysis of facial expressions in the wild. 
Nevertheless, convolution-based FER algorithms cannot consider the global information of the image due to the limitation of convolutional local receptive field. Influenced by the vision transformer, 
Xue et al. \cite{xue2021transfer} designed the first transformer-based FER network to model long-range dependencies for FER. 
Kim et al. \cite{kim2022facial} improved the vision transformer (ViT) to combine both global and local features so that ViT can be adapted to FER task.

Among many excellent FER works, POSTER \cite{zheng2022poster} stands out with state-of-the-art (SOTA) performance. 
POSTER mainly solves three key issues of FER at the same time: inter-class similarity, intra-class discrepancy and scale sensitivity. 
POSTER cleverly combines facial landmark with image features through a network design of two-stream pyramidal cross-fusion transformer.
With the difference and sparsity of landmark, POSTER successfully solves the issue of inter-class similarity and intra-class discrepancy in FER. 
The network design of pyramid architecture introduces multi-scale features for POSTER to solve the scale sensitivity problem. 
Along with the solution of the three main issues of FER, POSTER shows the amazing expression analysis ability.

Although POSTER works so well on FER, the huge number of parameters and expensive computational cost brought by its network architecture affects the efficiency of FER. 
To address this issue, we revisit the network design of POSTER and improve it to obtain POSTER++. 
We mainly improve POSTER in three directions: two-stream, cross-fusion and multi-scale feature extraction. 
POSTER contains two main branches: image-to-landmark and landmark-to-image. Landmark-to-image branch is essential as the core of POSTER to solve inter-class similarity and intra-class discrepancy. 
The image-to-landmark branch is only used to provide information to landmark that it fails to take into account. 
This does not contribute to solving the three main issues of FER. Therefore, in POSTER++, we remove the image-to-landmark branch from the two-stream design. 
This greatly reduces the computational cost on POSTER. For cross-fusion, we use a window-based cross-attention mechanism instead of the vanilla cross-attention mechanism in POSTER. 
The window-based cross-attention mechanism not only provides linear computational complexity for POSTER++ but also enhances the local modeling capability of the network. 
In addition, POSTER++ no longer uses an additional pyramid architecture for multi-scale feature extraction. We perform multi-scale feature extraction directly from the image backbone as well as from the facial landmark detector. 
For the extracted multi-scale features, we use a vision transformer network consisting of only two layers of transformer modules for integration. 
Based on the above designs, our POSTER++ becomes a simpler and more powerful facial expression recognition network. 
It achieves SOTA performance on several standard FER datasets with only 8.4G floating point operations (FLOPs) and 43.7M parameters (Param). 
Figure~\ref{fig1} demonstrates the superiority of POSTER++.

Specially, POSTER++ reached 92.21\% on RAF-DB \cite{li2017reliable}, 67.49\% on AffectNet \cite{mollahosseini2017affectnet} (7 cls) and 63.77\% on AffecNet (8 cls), respectively. 
This is better than POSTER (RAF-DB with 92.05\%, AffectNet (7 cls) with 67.31\% and AffectNet (8 cls) with 63.34\%). And POSTER++ offers a smaller Param (43.7M vs. 71.8M) and FLOPs (8.4G vs. 15.7G). 
We hope that our work could contribute to the design of future FER models.

In general, we summarize the contributions of this paper as follows:

\begin{itemize}
  \vspace{-5pt}
\item[1)] We design POSTER++ by modifying POSTER from three perspectives: two-stream, cross-fusion and feature extraction. Compared with POSTER, POSTER++ is simpler and stronger.
\vspace{-5pt}
\item[2)] POSTER++ shows state-of-the-art performance on several standard FER datasets such as RAF-DB, AffecNet and CAER-S. This shows the powerful expression analysis capability of POSTER++.
\vspace{-5pt}
\item[3)] POSTER++ greatly reduces the FLOPs and Param of POSTER. Specifically, POSTER++ reduces 28.1M of Param and 7.3G of FLOPs. This greatly improves the computational efficiency of the model.
\end{itemize}

\begin{figure*}[t]
   \centering
   \includegraphics[width=1\linewidth]{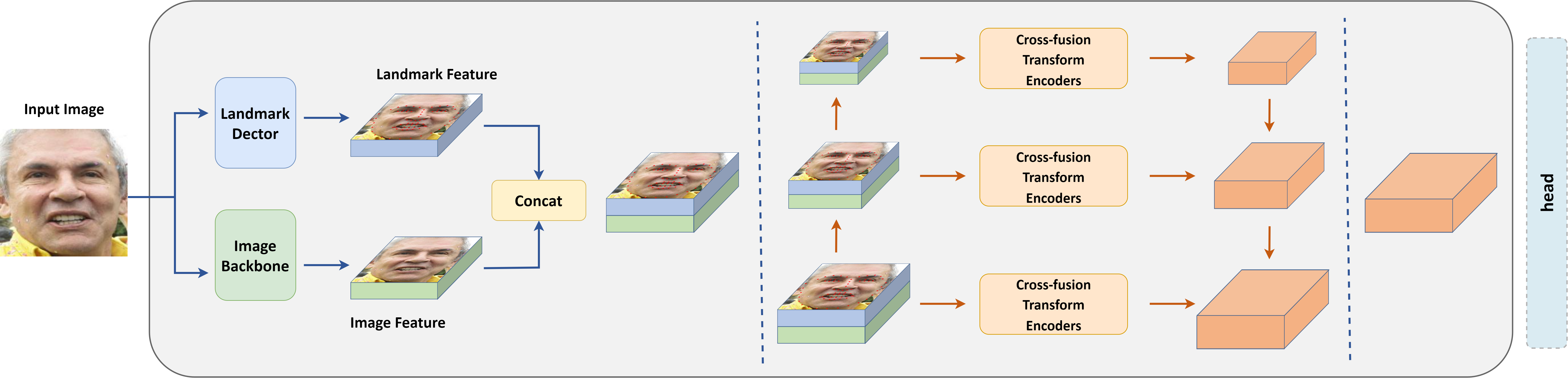}
    \caption{Pipeline of POSTER. POSTER mainly contains facial landmark detector, image backbone, cross-fusion transformer encoders and pyramid network.}
    \label{fig2}
 \end{figure*}

\section{Related Work}

\subsection{Facial Expression Recognition}

The study of FER has become very popular in recent years as more and more researchers focus on human-computer interaction. 
Zhao et al. \cite{zhao2007dynamic} used the manual feature LBP \cite{ojala2002multiresolution} for the research of FER with good results. 
Zhong et al. \cite{zhong2012learning} proposed a two-stage multitask sparse learning framework (MTSL) for the FER task by exploring some common and specific information among different expressions. 
Savchenko et al. \cite{savchenko2021facial} studied lightweight convolutional neural networks for FER task learning and verified the effectiveness of CNNs for FER. 
Sang et al. \cite{sang2018discriminative} focused on reducing intra-class variation in facial expression depth features and introduced a dense convolutional network \cite{huang2017densely} for the FER task. 
PSR \cite{vo2020pyramid} solves the practical issues associated with individual wild images in FER in terms of pose, orientation and input resolution with its super-resolution pyramidal network architecture. 
Zhang et al. \cite{zhang2022learn} proposed an erasing attention consistency method to handle the noise-labeled facial expression recognition task that is more challenging than the conventional FER.

With the rise of transformer in the field of computer vision, many FER methods combined with transformer have emerged. 
The vision transformer was first used for the study of FER by Xue et al. \cite{xue2021transfer} and achieved state-of-the-art performance. 
VTFF \cite{ma2021facial} excels in dealing with facial expression recognition tasks in the wild by virtue of its feature fusion. 
Huang et al. designed the teacher-student model PIDViT \cite{huang2022pidvit} for modeling the probability distribution of frontal and multi-pose facial expressions, and solved the problem of pose change and occlusion in FER. 
Zhao et al. \cite{fan2020ma} combined global and local attention in order to address the two key issues of occlusion and pose change in FER. 
POSTER \cite{zheng2022poster} solves the intra-class discrepancy, inter-class similarity and scale sensitivity issues of FER in the same time by integrating image features with facial landmark features through two-stream, cross-fusion and pyramid design.

However, the huge computational cost of POSTER has prevented researchers from investigating further improvements in FER. 
To solve this issue, we improved the architecture of POSTER and proposed POSTER++, which is simpler and more powerful for FER tasks.

\subsection{Vision Transformer}

Recently vision transformer has been widely used for computer vision tasks on large scale datasets with its excellent ability to model long distance dependencies.

Dosovitskiy et al. \cite{dosovitskiy2020image} pioneered the introduction of transformer from the field of natural language processing to computer vision. 
Touvron et al. \cite{touvron2021training} used a teacher-student strategy to accelerate the training of transformer by distilling tokens. 
Zhou et al. \cite{zhou2021deepvit} found that the reason why the transformer quickly saturates at deeper levels is that the attention map becomes increasingly similar as the transformer goes deeper. 
Based on this observation, they proposed the Re-attention model to regenerate the attention map in order to enhance the diversity among layers at a small computational cost. 
Touvron et al. also designed CaiT \cite{touvron2021going}, a deep vision transformer for optimal image classification. 
To solve the issue that ViT is inferior to traditional ResNet \cite{he2016deep} on datasets without huge data size, Yuan et al. proposed T2T-ViT \cite{yuan2021tokens}. 
Besides, Hassani et al. proposed CCT \cite{hassani2021escaping} which uses convolution rather than patch embedding layer for self-attention processing. 
This introduces convolutional inductive bias for the transformer. Chen et al. proposed CrossViT \cite{chen2021crossvit}, which combines image patches of different sizes by dual branches to produce stronger image features. 
Heo et al. \cite{heo2021rethinking} also verified whether pooling layers bring advantages to ViT as they do in convolutional neural networks (CNNs). 
Liu et al. \cite{liu2021swin} reduced the attention mechanism from quadratic computational complexity to linear by window attention and the design of a shift window scheme. 
Graham et al. grafted CNN with Transformer to obtain LeViT \cite{graham2021levit} with higher accuracy and faster speed. 
Wu et al. have designed a new architecture called convolutional visual transformer CVT \cite{wu2021cvt}, which improves the performance and efficiency of ViT by introducing convolution into vision transformer to produce the better results of both designs. 
Chen et al. proposed a new architecture with a pyramidal structure and a novel region-to-local-attention vision transformer, RegionViT \cite{chen2021regionvit}. 
Wang et al. \cite{wang2021pyramid} introduced ViT into a CNN-like pyramid structure for intensive prediction tasks such as object detection and semantic segmentation.

The architectural design of these vision transformer efforts inspires our improvements for POSTER. 
This leads to a better trade-off between accuracy and computational complexity in FER with our POSTER++.

\section{Method}

In this section, we first review the POSTER process. 
We then describe the overall architecture of POSTER++ and discuss the specific details of POSTER++ in three directions: two-stream, cross-fusion, and multi-scale feature extraction.
 \begin{figure*}[t]
   \centering
   \includegraphics[width=0.9\linewidth]{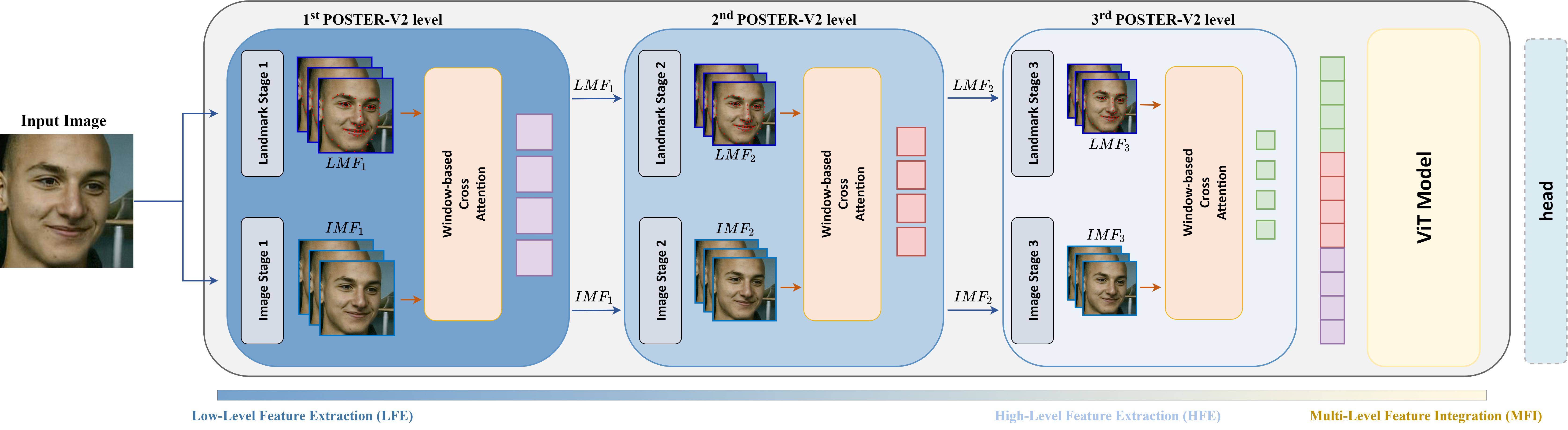}
    \caption{The overview of POSTER++ architecture. $LMF_i$ and $IMF_i$ denotes facial landmark features and image features at the i-th 	level of POSTER++ respectively.}
    \label{fig3}
 \end{figure*}
\subsection{A brief review of POSTER}

POSTER contains four main core designs: facial landmark detector, image backbone, cross-fusion transformer encoders and pyramid network. 
Given the input image ${X} \in {\mathbb{R}^{H \times W \times 3}}$, POSTER obtain the image features ${{X}_{{img}}}$ and landmark features ${{X}_{{lm}}}$ by facial landmark detector and image backbone, respectively.

The image features ${{X}_{img}} \in {\mathbb{R}^{N \times D}}$ as well as the landmark features ${{X}_{{lm}}} \in {\mathbb{R}^{N \times D}}$  are mapped into three matrices respectively: 
image query matrix ${Q}_{img}$, image key matrix ${K}_{img}$, image value matrix ${V}_{img}$ and landmark query matrix ${Q}_{lm}$, 
landmark key matrix ${K}_{lm}$, landmark value matrix ${{V}_{{lm}}}$ in the cross-fusion transformer encoder. Specifically expressed as:

\begin{equation}
   \begin{split}
         &{\text{Q}_{\text{img}}} = {\text{X}_{\text{img}}}{\text{W}_{\text{q1}}},{\text{Q}_{\text{lm}}} = {\text{X}_{\text{lm}}}{\text{W}_{\text{q2}}},\\
         &{\text{K}_{\text{img}}} = {\text{X}_{\text{img}}}{\text{W}_{\text{k1}}},{\text{K}_{\text{lm}}} = {\text{X}_{\text{lm}}}{\text{W}_{\text{k2}}},\\
         &{\text{V}_{\text{img}}} = {\text{X}_{\text{img}}}{\text{W}_{\text{v1}}},{\text{V}_{\text{lm}}} = {\text{X}_{\text{lm}}}{\text{W}_{\text{v2}}},
 \end{split}
   \label{eq1}
 \end{equation}
where ${{W}_{q1}},{W}_{q2},{W}_{k1},{W}_{k2},{W}_{v1}$ and ${{W}_{{v2}}} \in {\mathbb{R}^{D \times D}}$ are the mapping matrix.

The cross-fusion transformer encoder uses the vanilla cross-attention mechanism to interact image features and landmark features respectively. It is defined as follows:

\begin{equation}
   \begin{split}
      &\text{Attention}_\text{(img)} = {\text{softmax}}({{\text{Q}}_{{\text{lm}}}}{\text{K}}_{\text{img}}^{\text{T}}/\sqrt {\text{d}} ){{\text{V}}_{\text{img}}},\\
      &\text{Attention}_\text{(lm)} = {\text{softmax}}({{\text{Q}}_{{\text{img}}}}{\text{K}}_{\text{lm}}^{\text{T}}/\sqrt {\text{d}} ){{\text{V}}_{\text{lm}}},
\end{split}
   \label{eq2}
 \end{equation}
 where ${softmax}( \cdot )$ is softmax \cite{brown1992class} activation function and $\frac{1}{{\sqrt {d} }}$ is an appropriately normalized scaling factor used to prevent the gradient from being too small.

 In summary cross-fusion transformer encoder can be denoted as:

 \begin{equation}
   \begin{split}
         &\text{X'}_\text{img}=\text{Attention}_\text{(img)}+\text{X}_\text{img},\\
         &\text{X}_\text{img\_o}=\text{MLP(Norm(} \text{X'}_\text{img} \text{))}+\text{X'}_\text{img},\\
         &\text{X'}_\text{lm}=\text{Attention}_\text{(lm)}+\text{X}_\text{lm},\\
         &\text{X}_\text{lm\_o}=\text{MLP(Norm(} \text{X'}_\text{lm} \text{))}+\text{X'}_\text{lm},
\end{split}
   \label{eq3}
 \end{equation}

 where $MLP\left(  \cdot  \right)$ is multi-layer perceptron and $Norm\left(  \cdot  \right)$ denotes the normalization operation.

 Finally, POSTER extracts and integrates multi-scale features of images and landmarks by the pyramid network design. 
 The specific details are shown in Figure~\ref{fig2}.

 \subsection{Architecture}

 Figure~\ref{fig3} shows the pipeline for POSTER++. The POSTER++ keeps the facial landmark detector and image backbone in POSTER. 
 In difference, we remove the POSTER pyramid architecture and the image-to-landmark branch of the two-stream design. 
 Meanwhile, we perform multi-scale feature extraction directly from the facial landmark detector and image backbone.  
 And we introduce a small vision transformer consisting of only two layers of vanilla tranformer blocks in POSTER++ to integrate multi-scale features. 
 Moreover, we design the new cross-fusion transformer encoder with window-based cross-attention mechanism. 
 Next, we discuss the detailed modifications to POSTER++.

 \subsection{Two-stream}

 \begin{table}[h]
  \footnotesize
   \centering
     \begin{tabular}{l|c}
     Methods           & RAF-DB             \\
     \shline

     Baseline          & 91.00                 \\
     POSTER         & 92.05              \\
     POSTER w/o image to landmark branch & 91.82              \\
     POSTER w/o landmark to image branch & 91.62              \\
     \end{tabular}%
   \vskip 0.1in
     \caption{Ablation study of two branches in cross-fusion of POSTER. 
     The baseline in the table keeps the baseline setting in POSTER.}
   \label{tab1}
 \end{table}%

 Although two-stream is central to the design of POSTER, POSTER does not explore which branch of two-stream actually plays a major role. 
 Thus, in this section, we first perform an ablation study of the two-stream to learn the contribution of the two branches to the FER. 
 Table~\ref{tab1} shows the ablation results. We see that on the RAF-DB dataset, the accuracy of POSTER slips by 0.23 after missing the image to landmark branch. 
 If the landmark-to-image branch is missing, the accuracy of POSTER on RAF-DB is reduced by 0.43. 
 This indicates that although the image-to-landmark branch contributes to the POSTER FER performance, it is the landmark-to-image branch that plays a decisive role in POSTER. 
 Next, we analyze the above results at the theoretical level.

 \begin{figure}[h]
   \centering
   \includegraphics[width=1\linewidth]{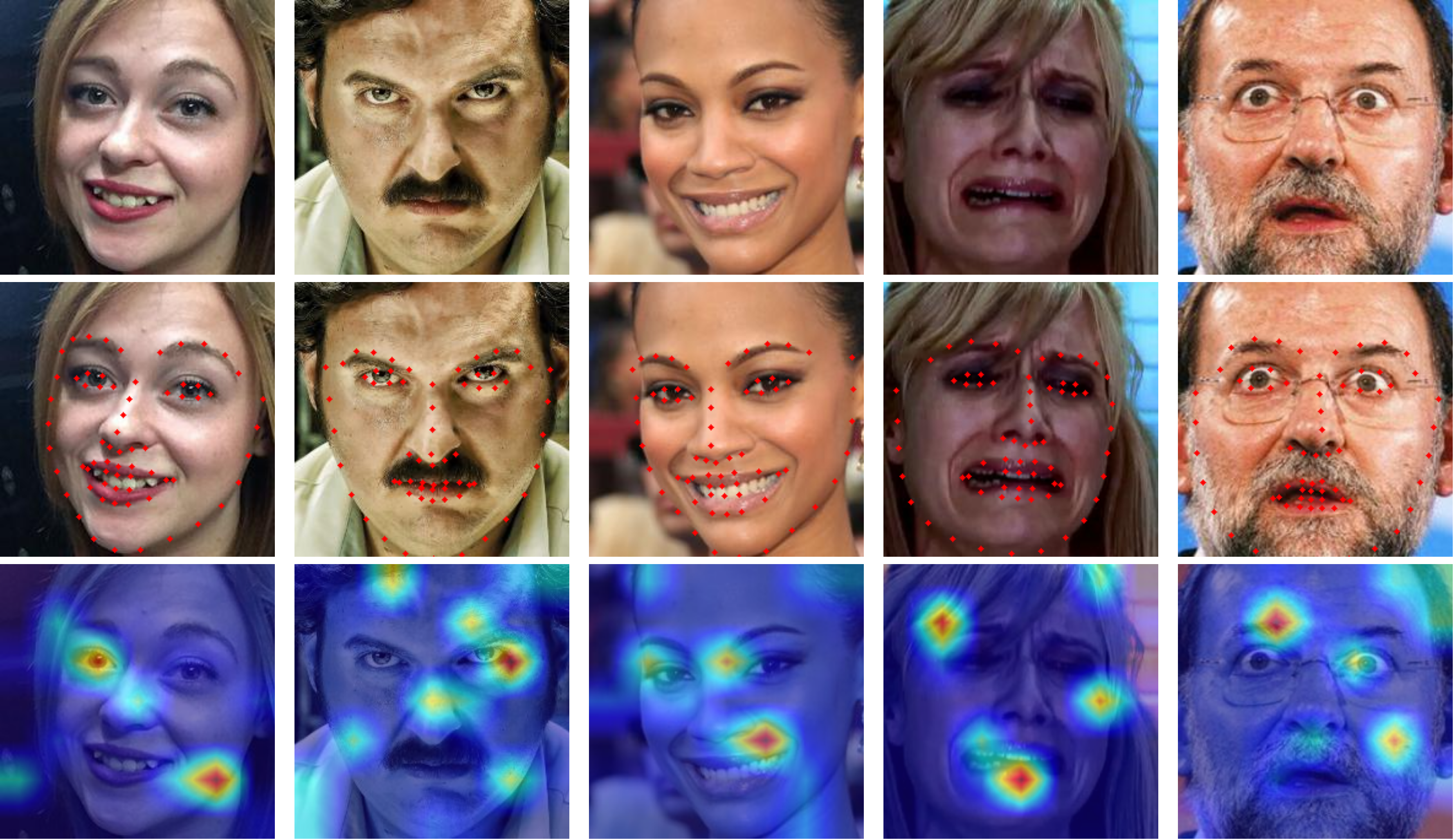}
    \caption{Input images (row \textcolor{red}{1}), facial landmark images (row \textcolor{red}{2}), landmark-to-image branching attention visualization results (row \textcolor{red}{3}). 
    We visualize the attention map belonging to the last layer of the landmarks to image branching for high-level features in POSTER. 
    We can observe that with the help of landmark features, the attention map focuses more on the outstanding areas of face and less on the areas common to face.}
    \label{fig4}
 \end{figure}

 \paragraph{Discussion.} The two-stream design in POSTER is mainly used to solve the issues of intra-class discrepancy and inter-class similarity in FER. 
 It includes landmark-to-image and image-to-landmark branches. 
 We revisit the influence of the two branches on POSTER. 
 In the landmark-to-image branch, the landmark features interact with the image features as queries ${{Q}_{{lm}}}$ in the cross-attention mechanism. 
 Image features are guided by landmark features to more easily represent salient regions of facial expressions when dealing with intra-class discrepancy issue. 
 Also benefiting from the sparsity of landmark features, image features guided by landmark features reduce the focus on regions where faces are prevalent. 
 This helps to reduce the impact of inter-class similarity in FER. The results of the visualization of landmark-to-image branch attention in Figure~\ref{fig4} also validate the above statements. 
 Therefore, the landmark-to-image branch in the two-stream is essential and needs to be retained. 
 For the image-to-landmark branch, the image features interact with the landmark features as query ${{Q}_{{img}}}$ to compensate for the lack of landmark features. 
 Although this also benefits the FER task to some extent, it does not contribute to solving the issues of inter-class similarity and intra-class discrepancy as well as comes with a huge computational cost. 
 As POSTER++ aims to solve the main POSTER FER problem while reducing its computational consumption.
 Thus, by making a trade-off between computational cost and accuracy of image-to-landmark branch, we eventually remove the image-to-landmark branch in the two-stream design.

 \subsection{Cross-fusion}

 \begin{figure}[h]
   \centering
   \includegraphics[width=0.75\linewidth]{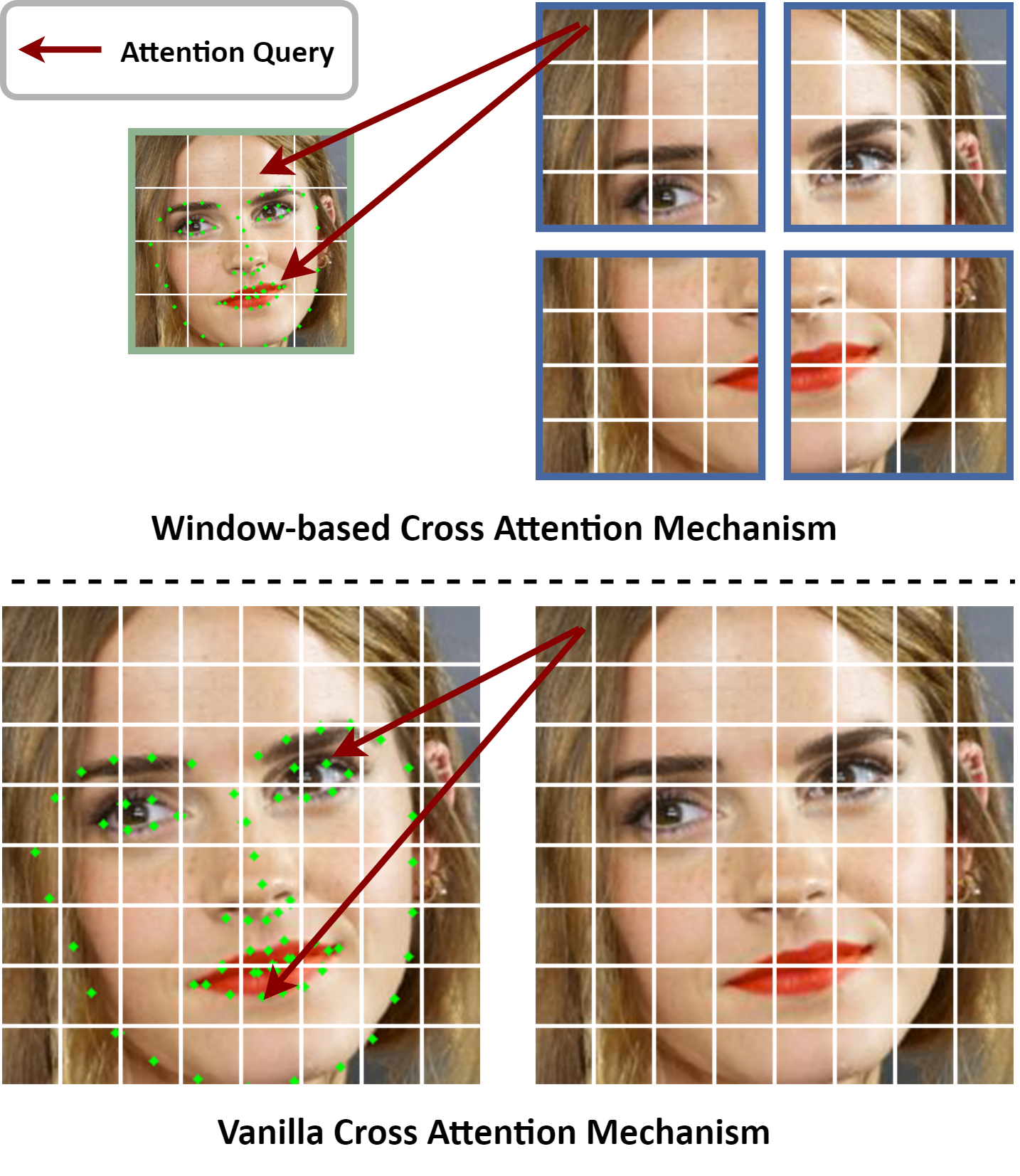}
    \caption{Window-based cross attention mechanism and vanilla cross attention mechanism.}
    \label{fig5}
 \end{figure}

 In POSTER++ we use window-based cross-attention mechanism instead of vanilla cross-attention mechanism in POSTER for the purpose of linear computation. 
 Figure~\ref{fig5} illustrates the detailed differences between the two cross-attention mechanisms. 
 For image features ${X_{img}} \in {\mathbb{R}^{N \times D}}$, we first divide them into several non-overlapping windows ${z_{img}} \in {\mathbb{R}^{M \times D}}$,
 where $z_{img}$ contains $M$ tokens.
 For the landmark feature ${{X}_{{lm}}} \in {\mathbb{R}^{C \times H \times W}}$, we first down-sample it to the window size ${{z}_{{lm}}} \in {\mathbb{R}^{c \times h \times w}}$,
 where $c = D,M = h \times w$.
 Then we reshape it according to the shape of ${Z}_{img}$. 
 At this point, the cross-attention with $I$ heads in a local window can be formulated as:

 \begin{equation}
   \begin{split}
      &\text{q}=\text{z}_\text{lm}\text{w}_\text{q},\text{k}=\text{z}_\text{img}\text{w}_\text{k},\text{v}=\text{z}_\text{img}\text{w}_\text{v},\\
      &\text{o}^\text{(i)}=\theta{\text{(q}^\text{(i)}}\text{k}^\text{(i)T}/\sqrt{\text{d}}+\text{b)v}^\text{(i)},\text{i}=\text{1,\dots,I},\\
      &\text{o}=\text{[}\text{o}^\text{(1)},\dots,\text{o}^\text{(I)}\text{]}\text{w}_\text{o},
\end{split}
   \label{eq4}
 \end{equation}
 where ${w}_{q},{w}_{k},{w}_{v},{w}_{o}$ are the mapping matrix, respectively. 
 $\theta\left( \cdot \right)$ is the softmax function.
 $\left[  \cdot  \right]$ denotes the merge operation and $b \in {\mathbb{R}^{I \times I}}$ is the relative position bias.

 We perform the above cross-attention calculation for all windows. 
 We refer to this cross-attention mechanism as window-based multi-head cross-attention (W-MCSA). 
 Thus the cross-fusion transformer encoder in POSTER++ can be expressed as follows:

 \begin{equation}
   \begin{split}
      &\text{X'}_\text{img}=\text{W-MCSA}_\text{(img)}+\text{X}_\text{img},\\
      &\text{X}_\text{img\_o}=\text{MLP(Norm(X'}_\text{img}\text{))}+\text{X'}_\text{img},
\end{split}
   \label{eq5}
 \end{equation}

 \paragraph{Computational Complexity Analysis.} Since the query in the two types of cross-attention computation keeps the same shape as the key, value, 
 we can use the multi-head self-attention and the window-based multi-head self-attention complexity to represent their computational complexity. 
 This can be indicated as follows:

 \begin{equation}
   \begin{split}
      &\Omega (\text{MCSA})=\text{4ND}^\text{2}+\text{2N}^\text{2}\text{D},\\
      &\Omega (\text{W-MCSA}) = \text{4ND}^\text{2}+\text{2M}^\text{2}\text{ND},
\end{split}
   \label{eq6}
 \end{equation}

 According to Eqn~\ref{eq6}, we can find that the window-based cross-attention mechanism we use successfully reduces the computational complexity of cross-fusion in POSTER from square level to linear level. 
 This further improves the computational efficiency of POSTER++.

 \subsection{Multi-scale feature extraction}

 From Figure~\ref{fig3}, we can observe that POSTER++ removes the pyramid design from POSTER. 
 Moreover, in POSTER++, we extract multi-scale features directly from facial landmark detector and image backbone. 
 And we also add a small vision transformer network to POSTER++ for the integration of multi-scale features. 
 For the obtained multi-scale features ${o^1},{o^2},{o^3}$, we directly merge in the token dimension and using the vanilla transformer blocks for processing. 
 This process is specifically described as:

 \begin{equation}
   \begin{split}
      &\text{o}=\text{[}\text{o}^\text{1},\text{o}^\text{2},\text{o}^\text{3}\text{]},\\
      &\text{o'}=\text{MSA(o)}+\text{o},\\
      &\text{o}_\text{out}=\text{MLP(Norm(o'))}+\text{o'},
\end{split}
   \label{eq7}
 \end{equation}
 where $MSA\left( \cdot \right)$ represents multi-head self-attention mechanism.
 For above design we discuss as follows.
 
 \paragraph{Discussion.} POSTER adopts the pyramid structure to solve the scale sensitivity problem in FER. 
 However, we consider that the pyramid structure design is only an up-sampling and down-sampling operation on the basis of the same scale feature map. 
 Although it provides multi-scale information to some extent, we believe that it is not as good as multi-scale feature extraction directly from the network. 
 For the integration of multi-scale features, we believe that the vanilla transformer block is sufficient for this task. 
 We combine the tokens of all scale feature maps together, and the attention mechanism can model long-range dependencies for all scale tokens. 
 Thus, different scales of token information are delivered in the transformer block. 

 \section{Experiments}

 We verify the effectiveness of POSTER++ on several standard FER datasets such as RAF-DB \cite{li2017reliable}, AffectNet \cite{mollahosseini2017affectnet} and CAER-S \cite{lee2019context}. 
 In the following, we first compare POSTER++ with SOTA methods. We then conduct a series of method analysis and ablation studies on POSTER++. 
 More detailed experimental setup, more experimental results and visualization results are detailed in the Appendix.

 \subsection{Experiment Setup}

 \begin{table}[h]
  \small
   \centering
     \begin{tabular}{l|ccc}
     Dataset           & Train size        & Test size         & Classes \\
     \shline
     RAF-DB            & 12271             & 3068              & 7 \\
     AffectNet (7 cls) & 280401            & 3500              & 7 \\
     AffectNet (8 cls) & 283501            & 4000              & 8 \\
     CAER-S            & 44996             & 20987             & 7 \\
     \end{tabular}%
   \vskip 0.1in
   \caption{Detailed size of the experimental dataset.}
   \label{tab2}%
 \end{table}%
 
 \paragraph{Datasets.} We evaluat the FER performance of POSTER++ on the widely used RAF-DB, AffectNet and CAER-S datasets. 
 The Real-world Affective Faces Database (RAF-DB) is a large-scale database of facial expressions, annotated by 315 staff members (students and faculty members of the University). 
 For the selection of expressions, RAF-DB selected six basic emotions as well as neutral emotions from a range of expressions 
 (e.g., smile, cackle, cry, anger, fear, dread, fear, shock, surprise, disgust, and no expression), for a total of seven expressions for expression annotation. 
 It mainly contains 12,271 training images as well as 3,068 test images. 
 AffectNet is currently the largest publicly available dataset in the FER field. It contains about 1M images of faces associated with emotional words. 
 It mainly contains 8 categories of primary emotions (neutral, happy, angry, sad, fear, surprise, disgust,contempt). 
 We mainly use AffectNet settings based on class 7 (excluding contempt) as well as class 8. AffectNet (7 cls) consists of 280K training images and 3.500 validation images (500 images per category). 
 AffectNet (8 cls) consists of 283K training images and 4.000 validation images (500 images per category). 
 The CAER-S dataset was obtained from the CAER dataset containing 65,983 images. 
 It is mainly divided into 7 types of expressions: neutral, happy, sad, surprised, fear, disgust and anger. 
 In the FER task we used 44996 images for training and 20987 images for testing. 
 The specific dataset configuration is shown in Table~\ref{tab2}.

 \begin{table}[h]
   \scriptsize
   \setlength{\tabcolsep}{1.0mm}
   \centering
     \begin{tabular}{l|c|c|c|c}
     Methods           & Year              & RAF-DB            & AffectNet (7 cls) & AffectNet (8 cls) \\
     \shline
     SCN \cite{wang2020suppressing}               & CVPR 2020         & 87.03             & -                 & 60.23 \\
     PSR \cite{vo2020pyramid}               & CVPR 2020         & 88.98             & 63.77             & 60.68 \\
     LDL-ALSG \cite{chen2020label}         & CVPR 2020         & 85.53             & 59.35             & - \\
     RAN \cite{wang2020region}              & TIP 2020          & 86.90              & -                 & - \\
     DACL \cite{farzaneh2021facial}             & WACV 2020         & 87.78             & 65.20              & - \\
     KTN \cite{li2021adaptively}              & TIP 2021          & 88.07             & 63.97             & - \\
     DMUE \cite{she2021dive}             & CVPR 2021         & 89.42             & 63.11             & - \\
     FDRL \cite{ruan2021feature}             & CVPR 2021         & 89.47             & -                 & - \\
     VTFF \cite{ma2021facial}             & TAC 2021          & 88.14             & 61.85             & - \\
     ARM \cite{shi2021learning}              & 2021              & 90.42             & 65.20              & 61.33 \\
     TransFER \cite{xue2021transfer}         & ICCV 2021         & 90.91             & 66.23             & - \\
     DAN \cite{wen2021distract}              & 2021              & 89.70              & 65.69             & 62.09 \\
     EfficientFace \cite{zhao2021robust}    & AAAI 2021         & 88.36             & 63.70              & 60.23 \\
     MA-Net \cite{zhao2021learning}           & TIP 2021          & 88.42             & 64.53             & 60.29 \\
     Meta-Face2Exp \cite{zeng2022face2exp}    & CVPR 2022         & 88.54             & 64.23             & - \\
     EAC \cite{zhang2022learn}              & ECCV 2022         & 90.35             & 65.32             & - \\
     POSTER \cite{zheng2022poster}        & 2022              & 92.05             & 67.31             & 63.34 \\
     \rowcolor{mygray}
     POSTER++         & -                 & 92.21             & 67.49             & 63.77 \\
     \end{tabular}%
   \vskip 0.1in
   \caption{Comparison results with SOTA FER algorithm on RAF-DB and AffectNet.}
   \label{tab3}%
 \end{table}%
 \begin{table*}[t]
  \footnotesize
   \centering
   \begin{tabular}{l|cccccccccc}
     Dataset           & Method            & Neutral           & Happy             & Sad               & Surprise          & Fear              & Disgust           & Anger             & Contempt          & mean Acc \\
     \shline
     RAF-DB            & POSTER         & \textcolor[rgb]{0.000, 0.690, 0.314}{92.35} & 96.96             & 91.21             & 90.27             & 67.57             & \textcolor[rgb]{0.000, 0.690, 0.314}{75.00} & \textcolor[rgb]{0.000, 0.690, 0.314}{88.89} & -                 & \textbf{86.04} \\
     RAF-DB            & POSTER++         & 92.06             & \textcolor[rgb]{0.000, 0.690, 0.314}{97.22} & \textcolor[rgb]{0.000, 0.690, 0.314}{92.89} & \textcolor[rgb]{0.000, 0.690, 0.314}{90.58} & \textcolor[rgb]{0.000, 0.690, 0.314}{68.92} & 71.88             & 88.27             & -                 & 85.97 \\
     \shline
     AffectNet (7 cls) & POSTER         & \textcolor[rgb]{0.000, 0.000, 1.000}{67.20} & 89.00                & 67.00                & 64.00                & \textcolor[rgb]{0.000, 0.000, 1.000}{64.80} & \textcolor[rgb]{0.000, 0.000, 1.000}{56.00} & 62.60              & -                 & 67.23 \\
     AffectNet (7 cls) & POSTER++         & 65.40              & \textcolor[rgb]{0.000, 0.000, 1.000}{89.40} & \textcolor[rgb]{0.000, 0.000, 1.000}{68.00} & \textcolor[rgb]{0.000, 0.000, 1.000}{66.00} & 64.20              & 54.40              & \textcolor[rgb]{0.000, 0.000, 1.000}{65.00} & -                 & \textbf{67.45} \\
     \shline
     AffectNet (8 cls) & POSTER         & 59.40              & \textcolor[rgb]{1.000, 0.000, 0.000}{80.20} & 66.60              & 63.60              & \textcolor[rgb]{1.000, 0.000, 0.000}{63.60} & \textcolor[rgb]{1.000, 0.000, 0.000}{59.80} & 58.80              & 54.71             & 63.34 \\
     AffectNet (8 cls) & POSTER++         & \textcolor[rgb]{1.000, 0.000, 0.000}{60.60} & 76.40              & \textcolor[rgb]{1.000, 0.000, 0.000}{66.80} & \textcolor[rgb]{1.000, 0.000, 0.000}{65.60} & 63.00                & 58.00                & \textcolor[rgb]{1.000, 0.000, 0.000}{60.20} & \textcolor[rgb]{1.000, 0.000, 0.000}{59.52} & \textbf{63.76} \\
     \end{tabular}%
   \vskip 0.1in
     \caption{Class-wise accuracy of POSTER and POSTER++ on RAF-DB, AffectNet (7 cls), and AffectNet (8 cls) datasets.
     Green, blue and red mark the highest value of single category in RAF-DB, AffectNet (7 cls) and AffectNet (8 cls) respectively.}
   \label{tab4}%
   \end{table*}%
 \paragraph{Settings.} Similar to POSTER \cite{zheng2022poster}, we also use the ir50 \cite{deng2019arcface} network pre-trained on the Ms-Celeb-1M \cite{guo2016ms} dataset as the image backbone. 
 And MobileFaceNet \cite{PFL} with frozen weights is used as our facial landmark detector. We employ the Adam \cite{kingma2014adam} optimizer for 200 epochs training. 
 A training scheme with a batch size of 144, a learning rate of 3.5e-4 and a weight decay of 1e-4 was used. 
 We use random horizontal flipping and random erasing as our data augmentation methods. For the loss function, we choose the standard cross-entropy loss. 
 We eventually realized POSTER++ on a single NVIDIA RTX 3090 via Pytorch.

 \subsection{Comparison with SOTA FER Methods}

 \paragraph{Results on RAF-DB.} We compare POSTER++ with the SOTA FER algorithms in recent years on the RAF-DB datasets in Table~\ref{tab3}. 
 The experimental results show that POSTER++ exhibits SOTA performance on RAF-DB. 
 Compared with POSTER (92.05), POSTER++ improved by 0.16. +1.86 for POSTER++ over EAC (90.35), and +1.3 for POSTER++ over TransFER (90.91). 
 This shows the superiority of PSTER V2 on RAF-DB. Table~\ref{tab4} shows the comparison of POSTER++ with POSTER for RAF-DB individual classes and average accuracy. 
 Although POSTER++ outperformed POSTER in several categories, the average accuracy was slightly inferior to POSTER.

 \paragraph{Results on AffectNet.}  In Table~\ref{tab3}, we also conduct FER experiments on AffectNet (7 cls) as well as AffectNet (8 cls). 
 We observe that POSTER++ exhibits SOTA FER effect in both AffectNet (7 cls) and AffectNet (8 cls). 
 Compared with POSTER (67.31, 63.34), POSTER++ increases 0.18, 0.43 on AffectNet (7 cls) and AffectNet (8 cls), respectively. 
 On AffectNet (8 cls), POSTER++ is higher than DAN (62.09) by 1.68. 
 On AffectNet (7 cls), POSTER++ is greater than TransFER (66.23) with 1.26. 
 This demonstrates that POSTER++ can maintain excellent FER performance even on larger datasets. 
 Table~\ref{tab4} shows that POSTER++ exceeds POSTER for the majority of individual class accuracies in both AffectNet (7 cls) and AffectNet (8 cls). 
 As a result, POSTER++ achieves better average accuracy than POSTER on AffectNet.

 \begin{table}[h]
  \small
   \centering
     \begin{tabular}{l|c|c}
     Methods           & Year              & CAER-S \\
     \shline
     DSN \cite{fan2018video}              & ICML 2018         & 75.19 \\
     CAER-Net-S \cite{lee2019context}        & ICCV 2019         & 73.51 \\
     GRERN \cite{gao2021graph}            & IEEE Access 2020  & 81.31 \\
     EfficientFace \cite{zhao2021robust}    & AAAI 2021         & 85.87 \\
     MA-Net \cite{zhao2021learning}           & TIP 2021          & 88.42 \\
     GLAMOR-Net \cite{le2022global}       & NCA 2021          & 89.88 \\
     POSTER \cite{zheng2022poster}        & 2022              & 92.73 \\
     \rowcolor{mygray}
     POSTER++         & -                 & 93.00 \\
     \end{tabular}%
   \vskip 0.1in
   \caption{Comparison results with SOTA FER algorithm on CAER-S.}
   \label{tab5}%
 \end{table}%

 \paragraph{Results on CAER-S.}  We compare POSTER++ with SOTA FER methods of recent years on the CAER-S dataset. 
 Our POSTER++ in Table~\ref{tab5} performs extremely well on the CAER-S dataset. 
 Specifically, POSTER++ scored 92.98 on CAER-S. +0.27 for POSTER++ over POSTER (92.73). +3.12 for POSTER++ over GLAMOR-Net (89.88), and +4.58 for POSTER++ over MA-Net (88.42). +7.13 for POSTER++ over EfficientFace (85.87). 
 The excellent results on CAER-S prove that the success of POSTER++ is no accident. 
 It shows the powerful generalization ability of POSTER++.

 \subsection{FLOPs and Param Comparison}

 \begin{table}[h]
   \footnotesize
   \centering
     \begin{tabular}{l|c|c|c|c}
     Methods           & \#Param           & \#FLOPs           & RAF-DB            & AffectNet \\
     \shline
     DMUE \cite{she2021dive}          &78.4M              &13.4G              &89.42              &63.11  \\
     TransFER \cite{xue2021transfer}       &65.2M               &15.3G              &90.91              &66.23  \\
     \hline
     POSTER-T \cite{zheng2022poster}      & 52.2M             & 13.6G             & 91.36             & 66.87 \\
     POSTER-S \cite{zheng2022poster}      & 62.0M             & 14.7G             & 91.54             & 67.13 \\
     POSTER \cite{zheng2022poster}        & 71.8M             & 15.7G             & 92.05             & 67.31 \\
     \hline
     POSTER++         & 43.7M             & 8.4G              & 92.21             & 67.49 \\
     \end{tabular}%
   \vskip 0.1in
   \caption{Comparison of Param and FLOPs with POSTER.}
   \label{tab6}%
 \end{table}%

 From Table~\ref{tab6}, we can see that POSTER++ achieves better FER results with smaller Param and FLOPs than POSTER and other algorithms.
 Compared to POSTER-T, POSTER++ reduces 8.5M Param and 5.2G FLOPs, while increasing 0.85\% on RAF-DB and 0.62\% on AffectNet. 
 Compared to POSTER-S, POSTER++ reduces 18.3M Param and 6.3G FLOPs, while increasing 0.67\% on RAF-DB and 0.36\% on AffectNet. 
 Compared to POSTER, POSTER++ reduces 28.1M Param and 7.3G FLOPs, while increasing 0.16\% on RAF-DB and 0.18\% on AffectNet. 
 Therefore, POSTER++ would be a better choice for the FER task.

 \subsection{Method Analysis}

 In this sub-section, we present a method analysis for the small ViT model we used in POSTER++ on RAF-DB.

 \begin{figure}[h]
   \centering
   \includegraphics[width=1\linewidth]{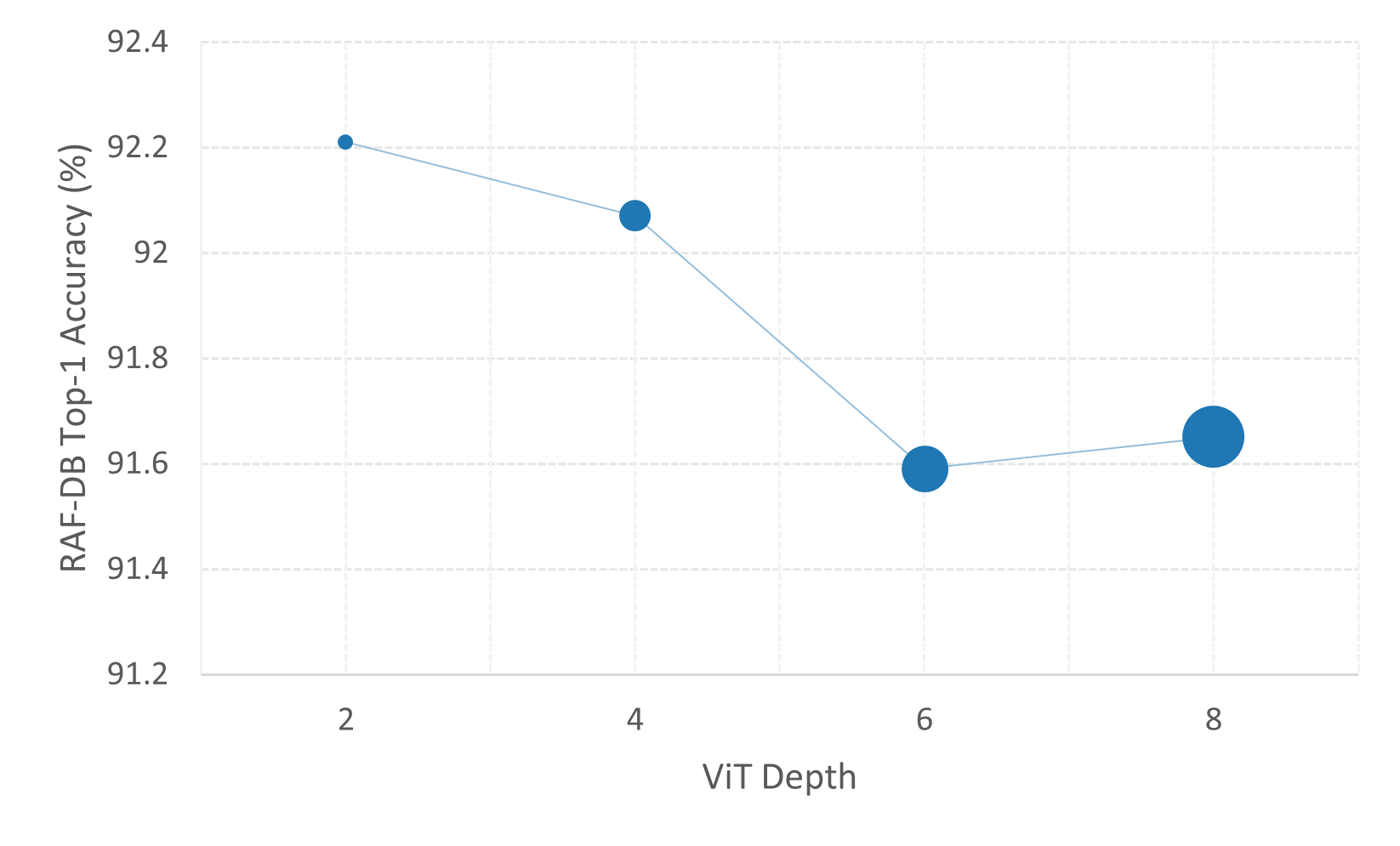}
    \caption{Influence of different depth ViT models on POSTER++ for RAF-DB.}
    \label{fig6}
 \end{figure}

 \paragraph{Vit depth.} Here, we investigate the impact of different depths for ViT on the FER performance of POSTER++. 
 In Figure~\ref{fig6}, we show the influence of the ViT model with depth \{2,4,6,8\} on POSTER++. 
 We observe that for multi-scale integration we do not need to increase the depth of the ViT model. 
 The ViT model with a depth of 2 is sufficient to handle the FER task. 
 A deeper ViT model hurts the performance of POSTER++ instead.

 \begin{table}[h]
  \small
   \centering
     \begin{tabular}{c|c|c}
     ViT w/ pre-trained weights & RAF-DB            & AffectNet  \\
     \shline
     \cmark                & 92.21             & 67.49 \\
     \xmark                & 91.49             & 60.20 \\
     \end{tabular}%
   \vskip 0.1in
   \caption{Impact of pre-trained ViT models for POSTER++ on FER.}
   \label{tab7}%
 \end{table}%
 \paragraph{Pre-trained Vit.}   We study the influence of the pre-trained ViT model on POSTER++. We use the ViT pre-trained weights on ImagenNet-21K \cite{ridnik2021imagenet} for POSTER++. 
 Table~\ref{tab7} shows that the performance of POSTER++ on FER drops after using the pre-trained ViT model. 
 We argue that this is mainly due to the fact that the pre-trained ViT model acts mainly on the feature extraction of the image-level inputs. 
 However, in POSTER++, ViT performs the multi-scale feature integration task of feature-level inputs. 
 The difference in input and task resulted in the pre-trained ViT not working on POSTER++.

 \subsection{Ablation Study}

 \begin{table}[h]
  \footnotesize
   \centering
     \begin{tabular}{l|c|c}
     Methods           & RAF-DB            & AffectNet \\
     \shline
     POSTER++         & 92.21             & 67.49 \\
     w/o multi-scale feature extraction & 91.47             & 66.51 \\
     w/o ViT           & 91.86             & 66.92 \\
     w/o W-MCSA        & 91.56             & 67.24 \\
     w/o cross-fusion  & 91.39             & 66.35 \\
     \end{tabular}%
   \vskip 0.1in
   \caption{Results of ablation experiments of key components of POSTER++.}
   \label{tab8}%
 \end{table}%
 
 We validate the effectiveness of our POSTER improvement component on the RAF-DB as well as on the AffectNet dataset.

 \paragraph{Multi-scale feature extraction.} We first verify the effectiveness of extracting multi-scale features directly in the network. 
 In this ablation experiment, we only use the image backbone as well as the last layer of feature maps from the facial landmark detector for cross-fusion. 
 From Table~\ref{tab8} we observe that POSTER++ degrades significantly on the RAF-DB and AffectNet datasets when multi-scale feature extraction is not performed. 
 This shows that our method of directly extracting multi-scale features can also solve the scale sensitivity issue of FER. 
 Also, this indicates the importance of multi-scale features for FER.

 \paragraph{Vit.} For the ViT used for multi-scale feature integration, we ablate it. 
 We directly sum several different scale features for FER. 
 According to the experimental results in Table~\ref{tab8}, POSTER++ decreases by 0.35 on RAF-DB and 0.57 on AffectNet when multi-scale feature integration is not performed by ViT. 
 This suggests that ViT facilitates multi-scale feature integration.

 \paragraph{W-MCSA.} We validate the effectiveness of W-MCSA for cross-fusion by ablation experiments. 
 In this experiment, we use the vanilla cross-attention mechanism to replace our window-based cross-attention mechanism. 
 We observed that POSTER++ degraded on both RAF-DB and AffectNet datasets. 
 This shows that the W-MCSA we use both improves the FER accuracy and reduces the computational complexity of POSTER. 
 Thus, W-MCSA is essential for POSTER++.
 
 \paragraph{Cross-fusion.} This experiment mainly verifies the role of landmark-to-image branch for POSTER++. 
 In the ablation experiments on cross-fusion, we merge the extracted image multi-scale features and landmark multi-scale features directly and integrate them by ViT. 
 Table~\ref{tab8} shows that the effectiveness of POSTER++ on RAF-DB as well as AffectNet drops sharply when cross-fusion is not applied. 
 This shows that cross-fusion is the key for POSTER++ to achieve SOTA FER. 
 Also, this indicates that addressing inter-class similarity and intra-class discrepancy are particularly important for FER task.

 \section{Conclusion}
 In this paper, we improve POSTER from three directions: two-stream, cross-fusion, and multi-scale feature extraction to obtain a simpler and stronger vision transformer for FER, POSTER++. 
 Extensive FER experimental results show that POSTER++ achieves the state-of-the-art FER performance while greatly reducing the Param and FLOPs of POSTER. 
 This suggests that POSTER++ achieves a better trade-off between accuracy and computational complexity. 
 Therefore, POSTER++ is a better choice for the FER task. 

 \section*{Acknowledge} This work was supported by Public-welfare Technology Application Research of Zhejiang Province in China under Grant LGG22F020032, and Key Research and Development Project of Zhejiang Province in China under Grant 2021C03137.

{\small
\bibliographystyle{ieee_fullname}
\bibliography{egbib}
}

\clearpage
\appendix
{\LARGE\noindent\textbf{Appendix}}

\section{Implementation Details}

For POSTER V2, we conduct FER experiments on the RAF-DB, AffectNet, and CAER-S datasets, respectively. 
For different datasets, we adopt different detail settings. 
Specifically, for different datasets, we exploit different learning rates for training according to the settings of POSTER V1. 
Moreover, for AffectNet (8 cls), POSTER V2 uses a classification head with a category number of 8 for prediction. 
The rest of the settings are consistent with the experimental sections in the main text.

\begin{table}[h]
   \tablestyle{6pt}{1.02}
   \footnotesize
   \centering
     \begin{tabular}{l|l}
     config            & value \\
     \shline
     optimizer         & Adam \\
     base learning rate & 3.50E-05 \\
     weight decay      & 1.00E-04 \\
     batch size        & 144 \\
     training epochs   & 200 \\
     learning rate schedule & ExponentialLR (gamma=0.98) \\
     \multirow{2}[2]{*}{augmentation} & RandomHorizontalFlip(), \\
                       & RandomErasing(scale=(0.02, 0.1)). \\
     drop path         & linspace(0, 0.5, 5) \\
     num classes       & 7 \\
     \end{tabular}%
   \vskip 0.1in
     \caption{Supervised training POSTER V2 from scratch on RAF-DB.}
   \label{tab9}%
 \end{table}%

\paragraph{RAF-DB Settings.}
We use the Adam optimizer with a learning rate of 3.5e-5 for 200 epochs training. 
The batch size is maintained at 144 and the weight decay remains at 1e-4. 
The learning rate schedule uses an exponential decay with a gamma of 0.98. 
Data augmentation includes random horizontal flipping and random erasure. 
The specific settings are shown in Table~\ref{tab9}.

\begin{table}[h]
   \tablestyle{6pt}{1.02}
   \footnotesize
   \centering
     \begin{tabular}{l|l}
     config            & value \\
     \shline
     optimizer         & Adam \\
     base learning rate & 1.00E-06 \\
     weight decay      & 1.00E-04 \\
     batch size        & 144 \\
     training epochs   & 200 \\
     learning rate schedule & ExponentialLR (gamma=0.98) \\
     \multirow{2}[0]{*}{augmentation} & RandomHorizontalFlip(), \\
                       & RandomErasing(p=1, scale=(0.05, 0.05)). \\
     drop path         & linspace(0, 0.5, 5) \\
     num classes       & 7 \\
     \end{tabular}%
   \vskip 0.1in
   \caption{Supervised training POSTER V2 from scratch on AffectNet (7 cls).}
   \label{tab10}%
 \end{table}%

\paragraph{AffectNet (7 cls) Settings.}
On the AffcetNet (7 cls) dataset, we adjust the learning rate to 1e-6. 
The training epochs remains at 200. The batch size is maintained at 144 and the weight decay remains at 1e-4. 
The learning rate schedule uses an exponential decay with a gamma of 0.98. 
Data augmentation includes random horizontal flipping and random erasure. 
The detailed settings are shown in Table~\ref{tab10}.

\begin{table}[h]
   \tablestyle{6pt}{1.02}
   \footnotesize
   \centering
     \begin{tabular}{l|l}
     config            & value \\
     \shline
     optimizer         & Adam \\
     base learning rate & 1.00E-06 \\
     weight decay      & 1.00E-04 \\
     batch size        & 144 \\
     training epochs   & 200 \\
     learning rate schedule & ExponentialLR (gamma=0.98) \\
     \multirow{2}[0]{*}{augmentation} & RandomHorizontalFlip(), \\
                       & RandomErasing(p=1, scale=(0.05, 0.05)). \\
     drop path         & linspace(0, 0.5, 5) \\
     num classes       & 8 \\
     \end{tabular}%
   \vskip 0.1in
   \caption{Supervised training POSTER V2 from scratch on AffectNet (8 cls).}
   \label{tab11}%
 \end{table}%

\paragraph{AffectNet (8 cls) Settings.}
We use the Adam optimizer with a learning rate of 1e-6 for 200 epochs training. 
The batch size is maintained at 144 and the weight decay remains at 1e-4. 
The learning rate schedule uses an exponential decay with a gamma of 0.98. 
Data augmentation includes random horizontal flipping and random erasure. 
In addition, we set the number of categories to 8. 
Table~\ref{tab11} shows the specific experimental settings.

\begin{table}[h]
   \tablestyle{6pt}{1.02}
   \footnotesize
   \centering
     \begin{tabular}{l|l}
     config            & value \\
     \shline
     optimizer         & Adam \\
     base learning rate & 4.00E-05 \\
     weight decay      & 1.00E-04 \\
     batch size        & 144 \\
     training epochs   & 200 \\
     learning rate schedule & ExponentialLR (gamma=0.98) \\
     \multirow{2}[0]{*}{augmentation} & RandomHorizontalFlip(), \\
                       & RandomErasing(p=1, scale=(0.05, 0.05)). \\
     drop path         & linspace(0, 0.5, 5) \\
     num classes       & 7 \\
     \end{tabular}%
   \vskip 0.1in
   \caption{Supervised training POSTER V2 from scratch on CAER-S.}
   \label{tab12}%
 \end{table}%

\paragraph{CAER-S Settings.}
On the CAER-S dataset, we employ the Adam optimizer with a learning rate of 4e-5 for 200 epochs of training. 
The batch size is maintained at 144 and the weight decay remains at 1e-4. 
The learning rate schedule uses an exponential decay with a gamma of 0.98. 
Data augmentation includes random horizontal flipping and random erasure. 
The specific settings are shown in Table~\ref{tab12}.

\section{Detailed Experimental Results}
\begin{figure*}[t]
   \centering
   \includegraphics[width=1\linewidth]{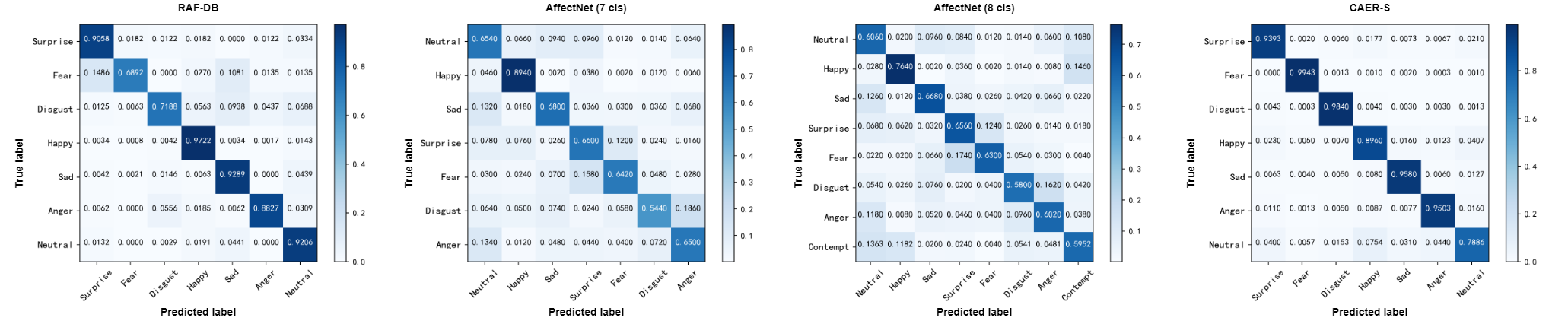}
    \caption{The confusion matrix of POSTER V2 on each dataset.}
    \label{fig7}
 \end{figure*}

 In this section, we show more detailed experimental results of POSTER V2 on each dataset. 
 And we also show the confusion matrix of POSTER V2 in each dataset in Figure~\ref{fig7}.

\begin{figure}[h]
   \centering
   \includegraphics[width=1\linewidth]{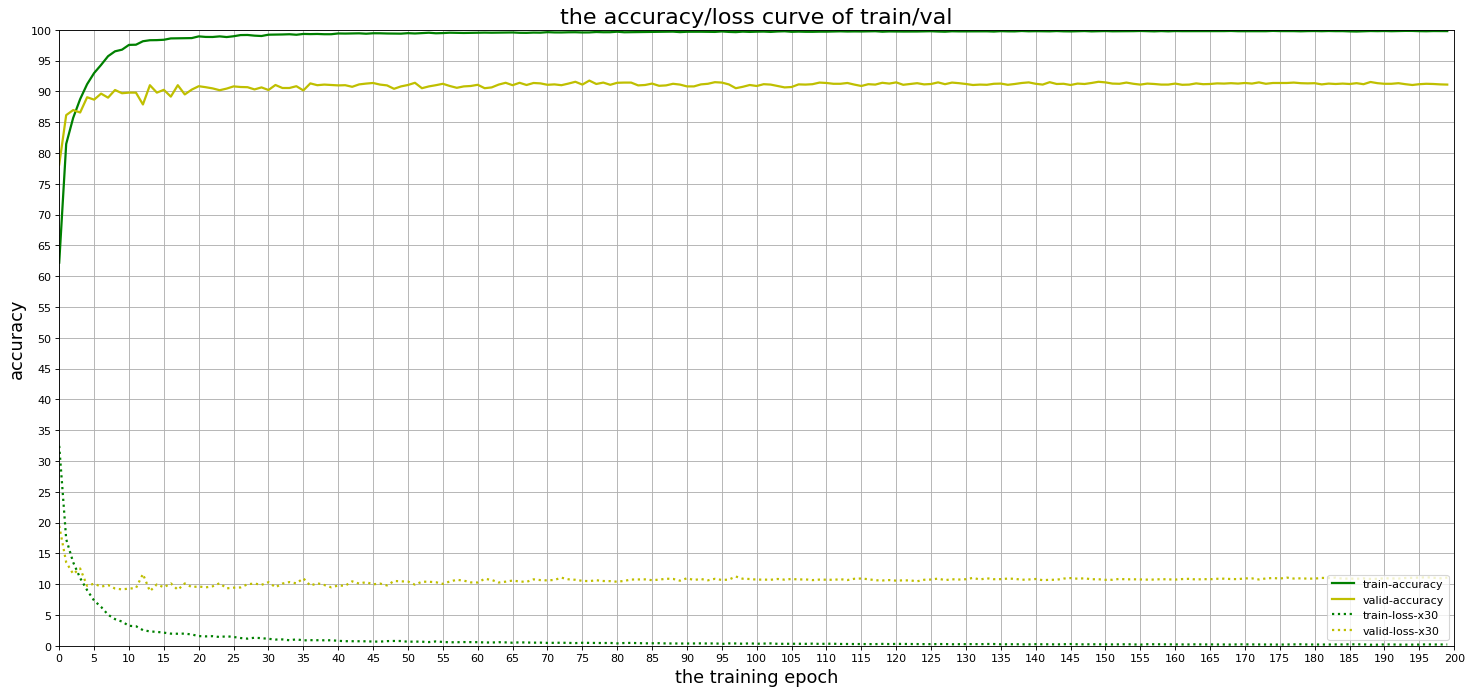}
    \caption{The specific training process of POSTER V2 on RAF-DB.}
    \label{fig8}
 \end{figure}

\paragraph{RAF-DB Results.}
Figure~\ref{fig8} shows the specific training process of POSTER V2 on RAF-DB. 
We observe that the training loss and validation loss of POSTER V2 decrease until saturation during the training process. 
Furthermore, the training accuracy and validation accuracy of POSTER V2 continue to increase until a small fluctuation.

 \begin{figure}[h]
   \centering
   \includegraphics[width=1\linewidth]{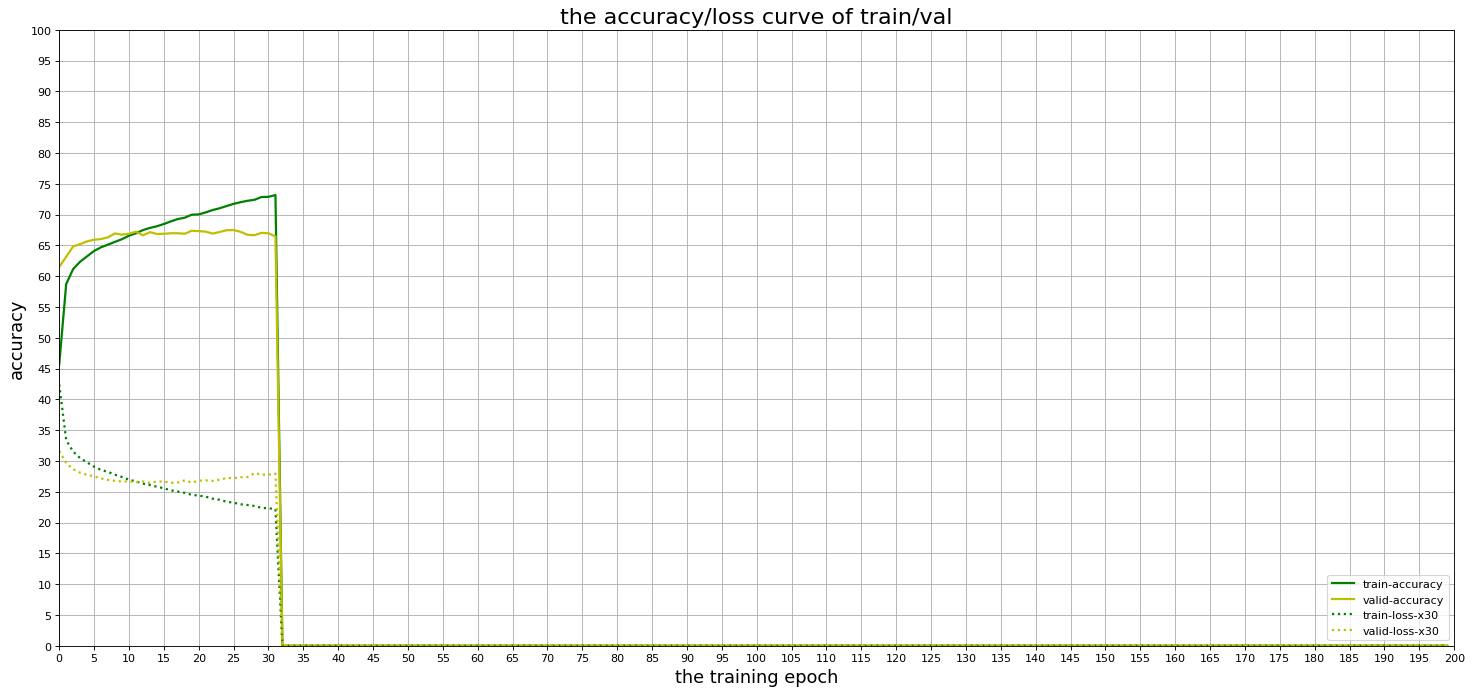}
    \caption{The detailed training process of POSTER V2 on AffectNet (7 cls).}
    \label{fig9}
 \end{figure}

 \paragraph{AffectNet (7 cls) Results.} We show in Figure~\ref{fig9} the detailed training of POSTER V2 on AffectNet (7 cls). 
 POSTER V2 achieves the best training results on AffectNet (7 cls) at an early stage. 
 At this point, POSTER V2 achieves the highest accuracy on AffectNet (7 cls) for both the training and test sets. 
 Therefore, we stop training in time to save training costs.

 \begin{figure}[h]
   \centering
   \includegraphics[width=1\linewidth]{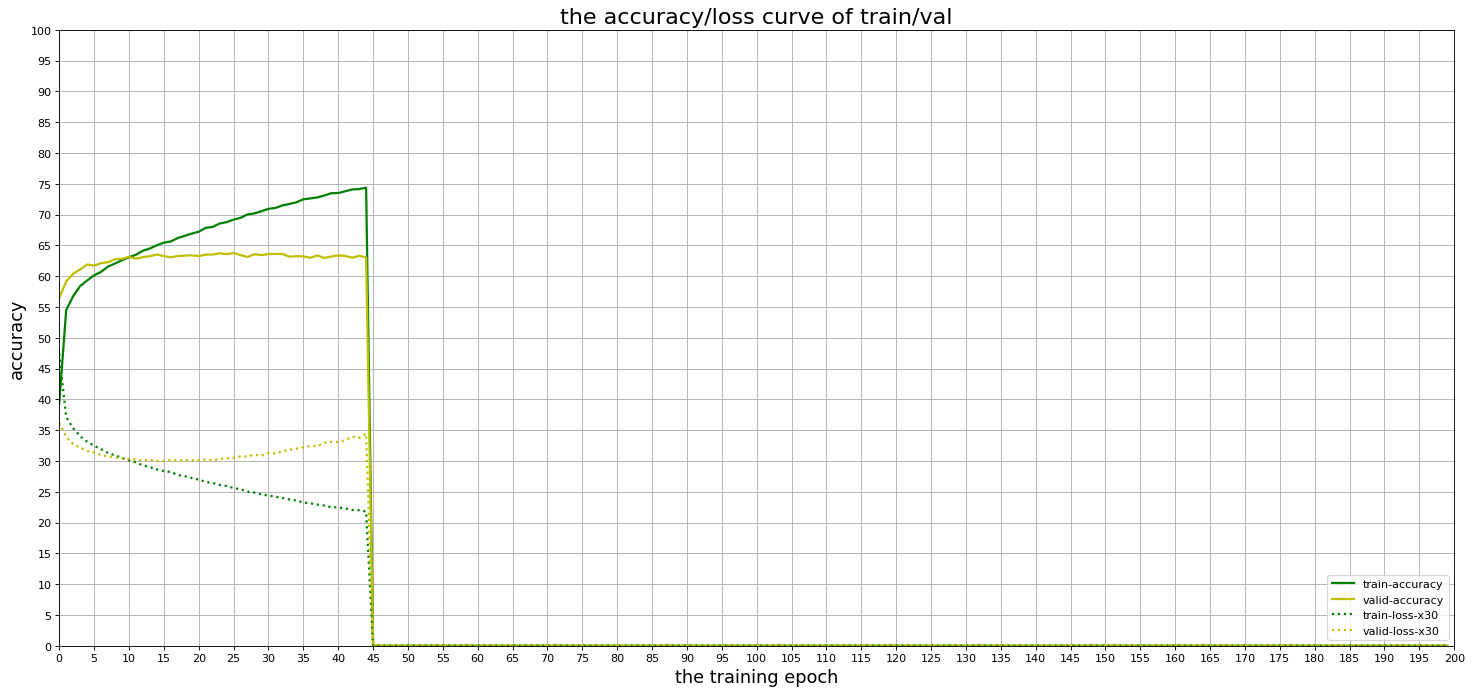}
    \caption{The detailed training process of POSTER V2 on AffectNet (8 cls).}
    \label{fig10}
 \end{figure}

 \paragraph{AffectNet (8 cls) Results.} Figure~\ref{fig10} shows the exact performance of POSTER V2 on AffectNet (8 cls). 
 We observe a similar phenomenon on AffectNet (8 cls) as POSTER V2 did on AffectNet (7 cls). 
 POSTER V2 also reach saturation in the early stages of AffectNet (8 cls). 
 POSTER V2 training loss continues to show a decreasing trend, yet there is a small increase in validation loss. 
 Nevertheless, the training accuracy of POSTER V2 on AffectNet (8 cls) continues to increase, and the validation accuracy has largely been optimal and remains constant. 
 Therefore, we take the same early end operation for POSTER V2 on AffectNet (8 cls) as we do for AffectNet (7 cls).

 \begin{figure}[h]
   \centering
   \includegraphics[width=1\linewidth]{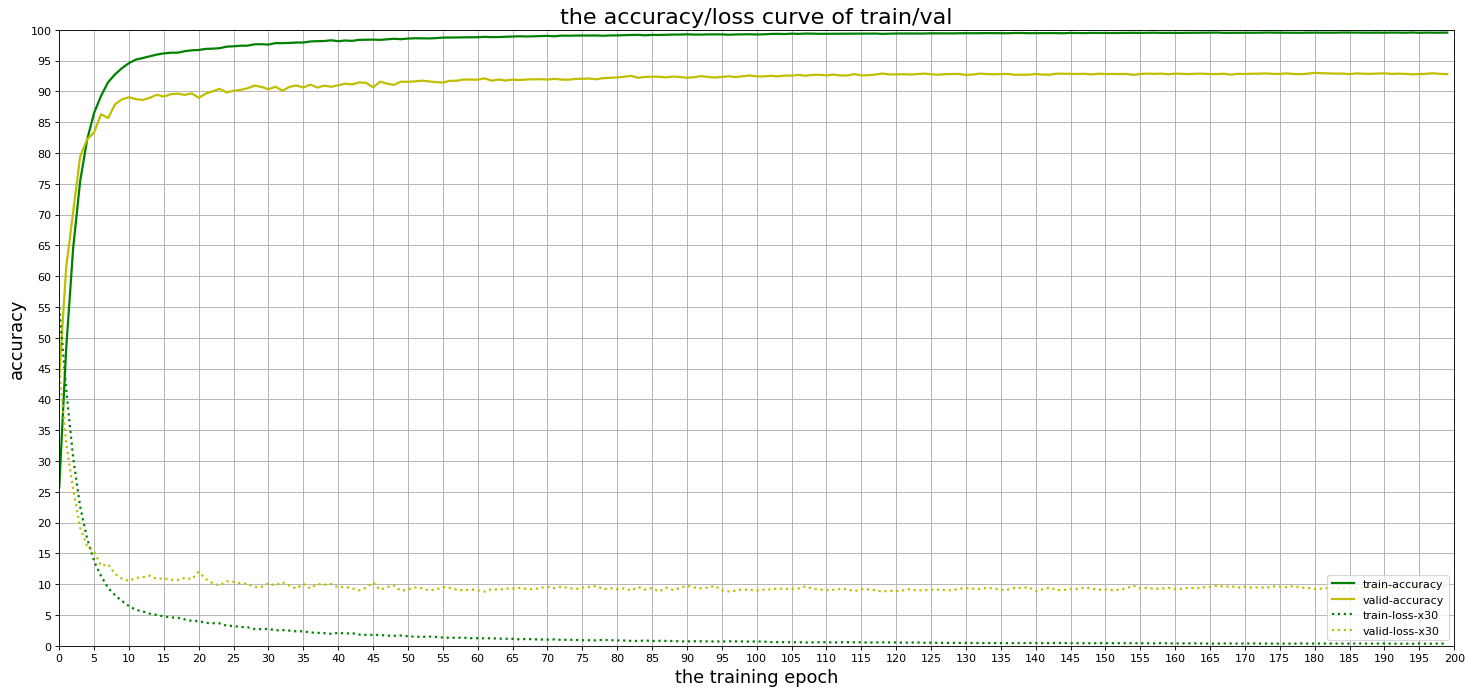}
    \caption{The detailed training process of POSTER V2 on CAER-S.}
    \label{fig11}
 \end{figure}

 \paragraph{CAER-S Results.} We show the specific training performance of POSTER V2 on CAER-S in Figure~\ref{fig11}. 
 Compared with other datasets, POSTER V2 has a relatively long saturation time on the CAER-S dataset. 
 During the training process, the loss on the POSTER V2 training and validation sets decreases and saturates at a late stage. 
 Meanwhile, the accuracy of POSTER V2 on both the training and validation sets has been increasing. 

\begin{figure*}[t]
  \centering
  \includegraphics[width=1\linewidth]{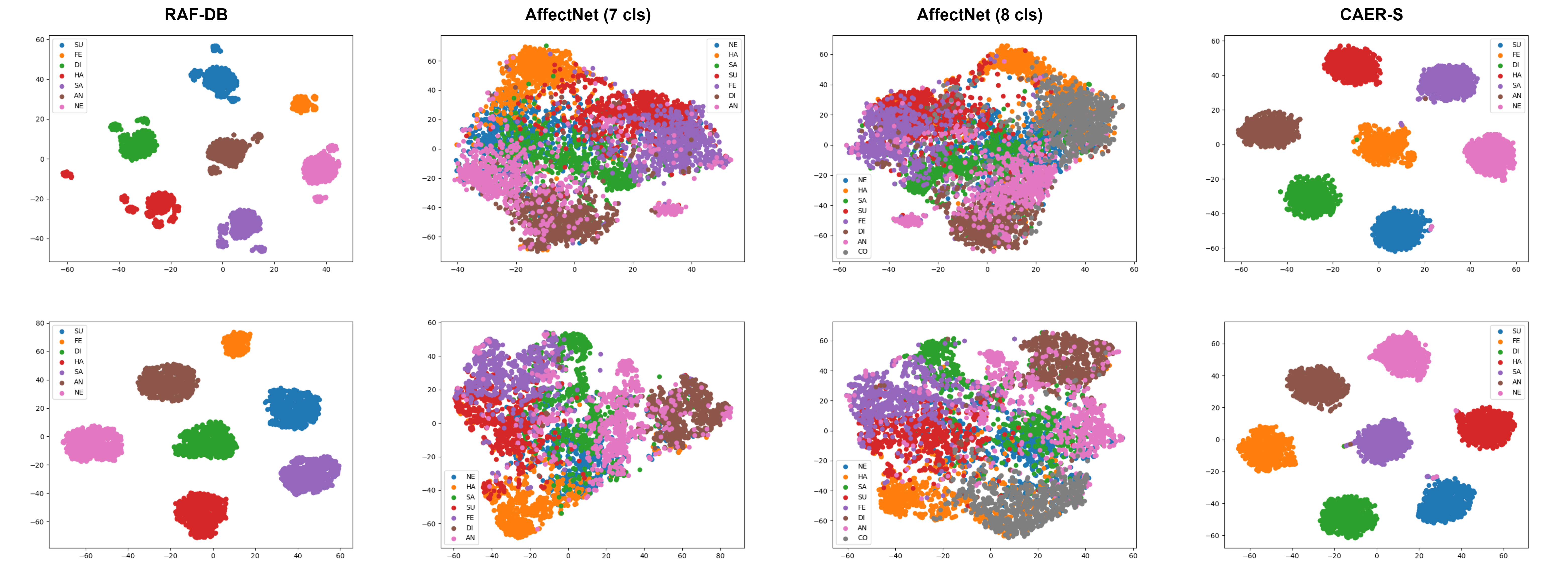}
   \caption{Comparison of POSTER V2 and POSTER V1 high-dimensional space t-SNE visualization results.
   POSTER V1 t-SNE visualization results (first row), POSTER V2 t-SNE visualization results (second row).}
   \label{fig12}
\end{figure*}

\section{Visualization}
\begin{figure*}[t]
  \centering
  \includegraphics[width=1\linewidth]{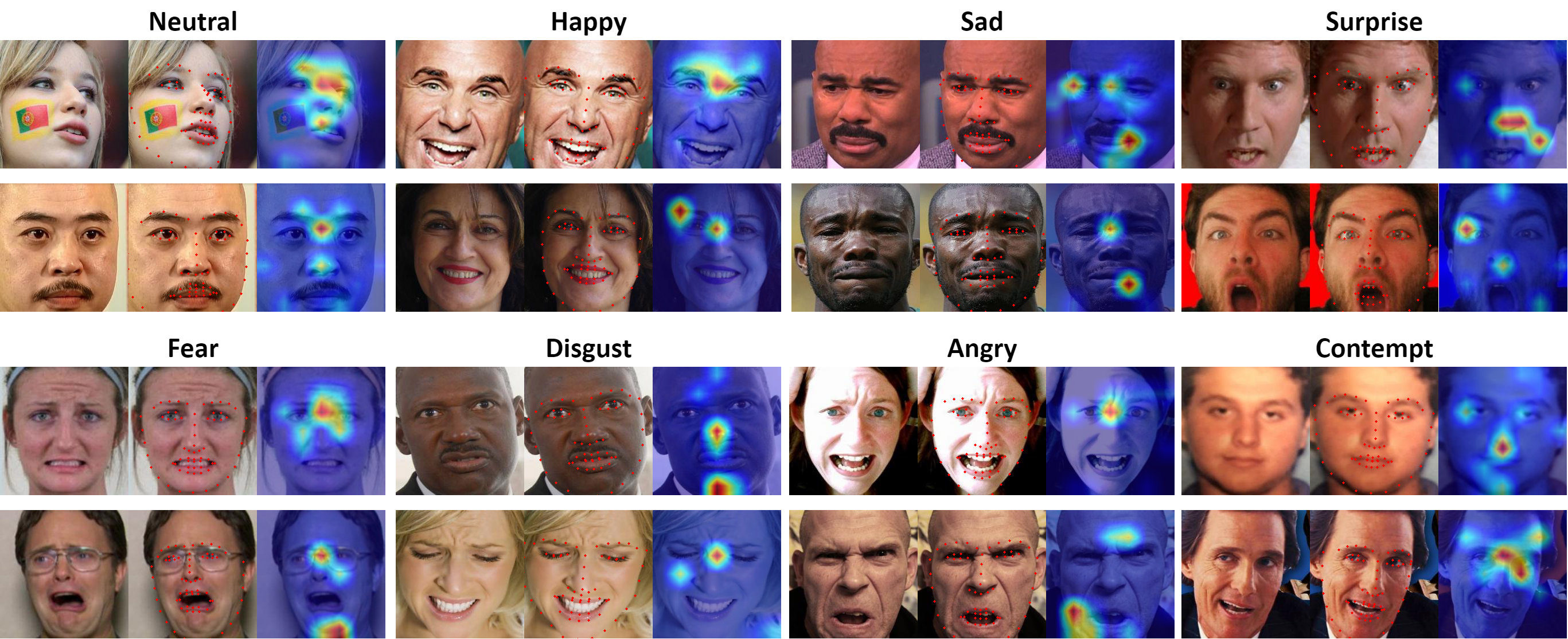}
   \caption{POSTER V2 cross-fusion stage attention visualization results. 
   For each triplet, we show the input image (left), the landmark image (middle), and attention map (right).}
   \label{fig13}
\end{figure*}

\paragraph{T-SNE Visualization.}
We visualized the high-dimensional features of POSTER V1 and POSTER V2 using t-SNE \cite{van2008visualizing}. 
As can be seen in Figure~\ref{fig12}, both POSTER V2 and POSTER V1 present good t-SNE visualization results on RAF-DB and CAER-S datasets. 
There is almost no significant difference between the t-SNE visualization results of POSTER V1 and POSTER V2 on CAER-S. 
POSTER V2 has a closer intra-class distance than POSTER V1 on RAF-DB. Although POSTER V1 and POSTER V2 have poor t-SNE visualization results on AffectNet (7 cls) and AffectNet (8 cls). 
But the inter-class distance between clusters in POSTER V2 is further than POSTER V1. 
Above results indicates that POSTER V2 is better than POSTER V1 in alleviating the issues of inter-class similarity and intra-class discrepancy in FER.

\paragraph{Attention Visualization.}
We visualize the attention map of the highest-level features of the POSTER V2 cross-fusion stage. 
From Figure~\ref{fig13}, we observe that POSTER V2 successfully captures important facial expression features with the help of facial landmark features.

\end{document}